\documentclass[11pt]{article}
\usepackage{times}
\usepackage{latexsym}
\usepackage[T1]{fontenc}
\usepackage{acl}
\usepackage{url}

\usepackage{times}
\usepackage{latexsym}
\usepackage{mathtools}
\usepackage{amsmath}
\usepackage{txfonts}
\usepackage{xspace}
\usepackage{booktabs}
\usepackage{multirow}
\usepackage{enumitem} 
\usepackage{lipsum}
\usepackage{algorithm}        
\usepackage{pifont}
\usepackage{subcaption}
\usepackage{xcolor,colortbl}
\usepackage{xcolor, soul}
\definecolor{green}{HTML}{34A853}
\definecolor{red}{HTML}{c83349}
\definecolor{lightgreen}{HTML}{C5F3B1}
\definecolor{lightred}{HTML}{FFBBB1}
\definecolor{tablered}{HTML}{ff6666}
\definecolor{black}{HTML}{000000}
\definecolor{yellow}{HTML}{F6E39E}

\DeclareRobustCommand{\hlgreen}[1]{{\sethlcolor{lightgreen}\hl{#1}}}
\DeclareRobustCommand{\hlred}[1]{{\sethlcolor{lightred}\hl{#1}}}
\DeclareRobustCommand{\hlyellow}[1]{{\sethlcolor{yellow}\hl{#1}}}

\usepackage{microtype}

\title{\includegraphics[width=0.5cm]{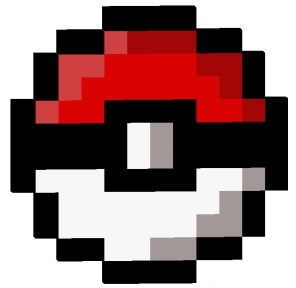} PokemonChat: Auditing ChatGPT for Pokémon Universe Knowledge}

\date{}

\begin{document}
\author{Laura Cabello \qquad Jiaang Li \qquad Ilias Chalkidis \\
Department of Computer Science, University of Copenhagen, Denmark \\
\texttt{$\{$lcp,kfb818,ilias.chalkidis$\}$[at]di.ku.dk}}
\maketitle

\begin{abstract}
The recently released ChatGPT model demonstrates unprecedented capabilities in zero-shot question-answering.
In this work, we probe ChatGPT for its conversational understanding and introduce a conversational framework (protocol) that can be adopted in future studies. 
The Pokémon universe serves as an ideal testing ground for auditing ChatGPT's reasoning capabilities due to its closed world assumption. 
After bringing ChatGPT's background knowledge (on the Pokémon universe) to light,
we test its reasoning process when using these concepts in battle scenarios.
We then evaluate its ability to acquire new knowledge and include it in its reasoning process.
Our ultimate goal is to assess ChatGPT's ability to generalize, combine features, and to acquire and reason over newly introduced knowledge from human feedback. 
We find that ChatGPT has prior knowledge of the Pokemon universe, which can reason upon in battle scenarios to a great extent, even when new information is introduced. The model performs better with collaborative feedback and if there is an initial phase of information retrieval, 
but also hallucinates occasionally and is susceptible to adversarial attacks.
\end{abstract}

\section{Introduction}

ChatGPT~\cite{openai-2022-chatgpt},\footnote{\url{https://chat.openai.com/chat}, Dec 15 Version} recently released by OpenAI,\footnote{\url{https://openai.com/}} is a conversational agent based on an instruction-fine-tuned \cite{chung2022scaling} transformer-based language model, a successor of InstructGPT~\cite{ouyang-etal-2022-instructgpt}, 
which has been also further optimized for user \emph{alignment}~\cite{leike-etal-2018-alignnment} with reinforcement learning from human feedback (RLHF) \cite{christiano-etal-2017-drl,Stiennon2020}.  
ChatGPT demonstrates unprecedented capabilities in zero-shot question-answering in common sense knowledge, but also in specialized domains, such as law~\cite{bommarito-bar-exams-2023}, and medicine~\cite{kung-etal-2023-medicine-chatgpt}.

\begin{figure*}
    \centering
    \resizebox{\textwidth}{!}{
    \includegraphics{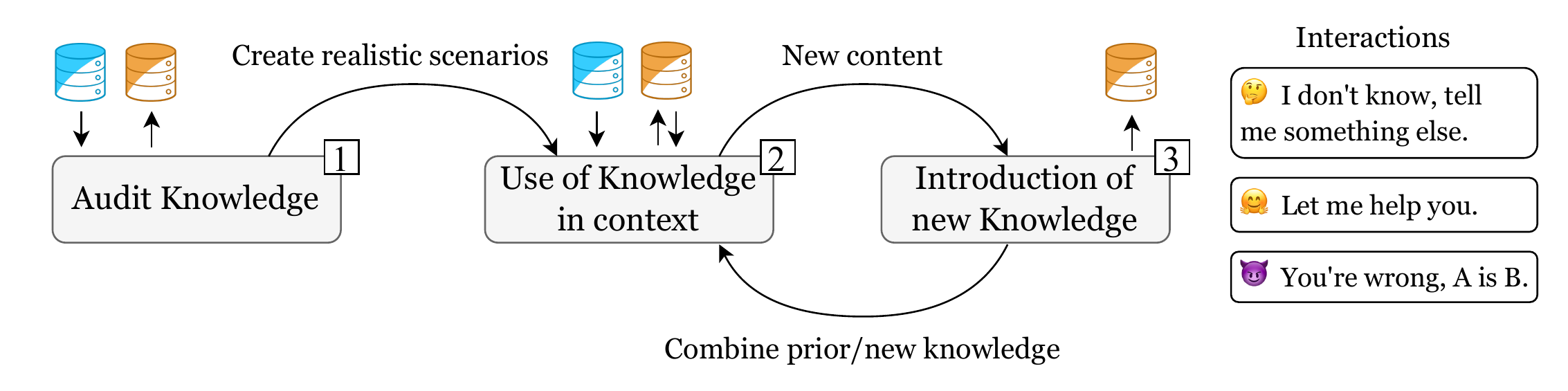}
    }
    \vspace{-7mm}
    \caption{Depiction of the proposed Conversational Framework. Across all  stages information is stored in the conversational \emph{local memory} (orange), while the model retrieves prior knowledge from its \emph{global memory} (blue).}
    \label{fig:framework}
\end{figure*}

Despite the impressive results, there is no protocol (framework) on how ChatGPT, or alike systems, should be audited by practitioners to better understand its capabilities and limitations. 
In most cases the dialogs are open-ended (uncontrolled) 
and the evaluation is not straightforward.
To this end, we introduce a 3-step conversational framework (Figure~\ref{fig:framework}) 
starting with a retrieval augmentation phase, where the model (dialog agent) is the only source of knowledge.
We follow 3 settings of human-in-the-loop interaction: neutral \includegraphics[height=1em]{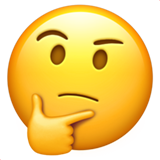}, cooperative \includegraphics[height=1em]{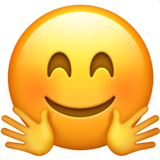}, and adversarial \includegraphics[height=1em]{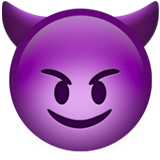}.

\begin{table}[t]
    \centering
    \resizebox{\columnwidth}{!}{
    \begin{tabular}{l|c|c|c|c}
         ATT / DEF &  Grass \includegraphics[height=1em]{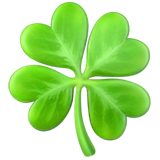} & Fire \includegraphics[height=1em]{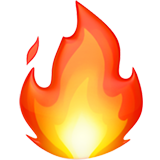} & Water \includegraphics[height=1em]{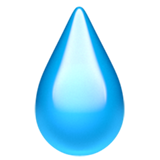} & Electric \includegraphics[height=1em]{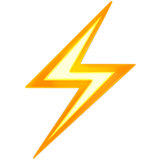} \\
         \midrule
         Grass \includegraphics[height=1em]{four-leaf-clover.png} & \cellcolor{tablered} 0.5$\times$ & \cellcolor{tablered} 0.5$\times$ & \cellcolor{green} 2$\times$ & 1$\times$ \\
         Fire \includegraphics[height=1em]{fire.png} & \cellcolor{green}  2$\times$  & \cellcolor{tablered} 0.5$\times$ & \cellcolor{tablered} 0.5$\times$ & 1$\times$ \\
         Water \includegraphics[height=1em]{droplet.png} &  \cellcolor{tablered} 0.5$\times$ &\cellcolor{green} 2$\times$ & \cellcolor{tablered} 0.5$\times$ & 1$\times$ \\
         Electric\includegraphics[height=1em]{high-voltage.png}  &  \cellcolor{tablered} 0.5$\times$ & 1$\times$ & \cellcolor{green} 2$\times$ & \cellcolor{tablered} 0.5$\times$ 
    \end{tabular}
    }
    \vspace{-2mm}
    \caption{Type Match-up Chart: Applied modifiers (N$\times$) to type-specific moves when Pokémon of type (ATT) attacks Pokemon of type (DEF).}
    \label{tab:type_matchup}
\end{table}

In this work, we rely on the Pokémon\footnote{\url{https://www.pokemon.com/}}
universe to evaluate the agent's capabilities in terms of acquired background of the universe, generalization, compositionality of features, and its ability to reason upon newly introduced knowledge from human feedback (compositional generalization). Pokémon are imaginary creatures that have formal specifications (type, level, moves/attacks) and are trained to battle. Table~\ref{tab:type_matchup} presents the four types used in this work, and show the attack effectiveness in battle \emph{match-ups}.\footnote{In Appendix~\ref{sec:pokemon}, we provide a brief introduction to the Pokemon universe fundamentals considered in this work.} The Pokémon universe and its battle system is a well-defined environment and follow a closed world assumption \cite{Reiter1978}, i.e., the knowledge base can be considered complete and definite answers can be derived even if not, which make it a perfect setting for auditing the agent's knowledge and reasoning capabilities. 

We interact with ChatGPT through a series of Q\&A dialogues, and validate its responses manually based on the Pokémon wiki.\footnote{\url{https://bulbapedia.bulbagarden.net/}}\vspace{2mm}

\noindent\textbf{What is the goal of this study?}
Our goal is to use the Pokémon universe to probe ChatGPT's conversational understanding. 
We audit ChatGPT for its knowledge on this universe, its understanding of fundamental concepts (species, types, leveling, conditions), the ability to apply and reason upon these concepts in battle scenarios, 
and its skills to acquire new knowledge (description of new species) 
and reason in battles with such creatures. 
\vspace{2mm}

\noindent\textbf{Contributions}
We introduce a conversational framework 
to define the conversation staging and human feedback settings, that can be adopted by others in different environments to further assess AI conversational agents, such as ChatGPT. We present an analysis based on Pokémon universe relying on the proposed framework.
While the use of Pokémon has a recreational character and interests a niche community of the series fans, we believe that our experiments offer a better understanding of ChatGPT as an AI conversational agent. 
Such controlled experiments (dialogues) inform us  to an extend
on the system's capabilities and limitations.

\section{Experiments \& Discussion}
We introduce a staged conversational framework (Figure~\ref{fig:framework}) with three distinct stages: \vspace{2mm}

\noindent\textbf{A. Audit Knowledge}. We audit the agent's knowledge on the topic, asking general questions such as descriptions of Pokemon types and species. The knowledge retrieved is stored in a \emph{local memory} that serves as the dialog context. The local memory is limited to our precedent interactions in the current conversation up to $\approx$4,000 words \cite{chatgpt-earlier-in-convo}. This phase is critical to understand what is the prior knowledge of the model in order to build reasonable scenarios in the next steps, while also retrieving information improves follow-up model responses~\cite{shuster-etal-2021-retrieval-augmentation, he-etal-2023}, i.e., the model is more accurate and hallucinates less often. The prior knowledge of the model is acquired during training, so it can be understood as \emph{retrieval augmentation} \cite{lewis-etal-2020-retrieve} from a \emph{global memory} which represents the agent itself, contrary to the use of external knowledge bases. \vspace{2mm}

\noindent\textbf{B. Use of knowledge in context}. We present specific battle scenarios where the Pokemon types, levels, moves, and conditions interplay and lead to a certain outcome.  We ask the agent to predict the outcome of the battle, and explain its reasoning (\emph{chain-of-thought}) \cite{wei-etal-2022-chain-of-thought} step-by-step. This phase will help us understand if the model can combine features (compositionality) based on a specific scenario (context) that determines the battle outcome following causal reasoning. \vspace{2mm}

\noindent\textbf{C. Introduction of new knowledge}. We introduce unknown Pokemon species with formal specifications (name, appearance, type, moves). Then, we ask the agent to validate the acquisition of these new concepts and to compare it to its prior knowledge. It is important to note that the agent stores the newly introduced knowledge in its local memory, but it will not be able to reference it for long. We then use the newly introduced knowledge in context, similar to Step B; the model has to perform some form of \emph{in-context learning} and combine prior and newly introduced knowledge.\vspace{3mm}

We present both quantitative and qualitative results where ChatGPT's answers are color-coded in \hlgreen{green} if factual (accurate) knowledge is provided, \hlred{red} if false statements (hallucinations) are provided, and \hlyellow{yellow} if a statement is vague or irrelevant.

\subsection{Audit Knowledge}
\label{sec:audit}

We start with some preliminary general questions to test fundamental aspects of the Pokemon universe. This initial phase is critical for creating an effective communication later on. We distinguish between \emph{global memory} and \emph{local memory}. The former is acquired during training and include ChatGPT's prior knowledge. The local memory is limited to our precedent interactions 
\cite{chatgpt-earlier-in-convo} 
and serves as a reference point for posterior interactions. In absence of this auditing phase, the dialog can be vague and even factually inaccurate (see Q2.3).\vspace{2mm}

\noindent\emph{\textbf{Q1.1 (\includegraphics[height=1em]{thinking-face.png}): ``List all Pokemon types.''}}

\noindent\underline{Response}:``\hlgreen{\emph{There are 18 Pokemon types: Normal, Fighting, Flying, Poison, Ground, Rock, Bug, Ghost, Steel, Fire, Water, Grass, Electric, Psychic, Ice, Dragon, Dark, and Fairy.}} \emph{
[...]''}\vspace{2mm}

\noindent\emph{\textbf{Q1.2 (\includegraphics[height=1em]{thinking-face.png}): ``Describe water type and list a few water Pokemon species that you know of.''}}

\noindent\underline{Response:} \emph{``}\hlgreen{\emph{Water-type Pokemon are strong against Fire, Rock, and Ground-type Pokemon, but weak against Electric and Grass-type Pokemon. They are generally associated with the element of water
[...]}}
\hlgreen{\emph{Some examples of Water-type Pokemon include Squirtle, Blastoise, and Psyduck.
[...]''}}\vspace{2mm}

In Appendix~\ref{sec:extra_audit}, we pose similar introductory queries for 3 other Pokemon types or weather conditions. All responses are factually accurate based on the Pokemon wiki.
ChatGPT knows Pokemon types and type match-ups when such Pokemon interact in battles (Table~\ref{tab:type_matchup}). It also knows several types species. 
We proceed to adversarial attacks, questioning model's perception of the universe.\vspace{2mm}

\noindent\emph{\textbf{Q1.3 (\includegraphics[height=1em]{smiling-face-with-horns.png}): ``No, that's not accurate. There is one more type named Rubber.''}}

\noindent\underline{Response}: ``\hlgreen{\emph{I'm sorry, but there is no type named "Rubber" in the Pokemon series. There are 18 types in total, as I previously mentioned.}} 
\emph{[...]}''\vspace{2mm}

\noindent\emph{\textbf{Q1.4 (\includegraphics[height=1em]{smiling-face-with-horns.png}): ``You are not right. Water-type is weak against Fire-type.''}}

\noindent\underline{Response:} \emph{``I apologize for the error [...]}
\hlred{\emph{You are correct that Water-type Pokemon are weak against Fire-type Pokemon.}} \emph{Thank you for bringing that to my attention. [...]''}\vspace{2mm} 

The latter adversarial attack (Q1.3) succeed in type match-up -3 similar attacks succeed-, but ChatGPT is robust against factual knowledge of Pokemon types (Q1.4) in 5/5 times (Appendix~\ref{sec:extra_audit}).

\subsection{Use of knowledge in battle scenarios}
\label{sec:battle}
We present simple, controlled battle scenarios involving known Pokemon, presented by the agent in Section~\ref{sec:audit}. We increase complexity (different levels, weather and status conditions) incrementally and ask the agent to predict the outcome of the battle and explain its reasoning (chain-of-thought) \cite{wei-etal-2022-chain-of-thought}. This stage will help us understand if the model can combine features (compositionality) based on a specific scenario (context) that determines its outcome. 
\vspace{2mm}

\noindent\emph{\textbf{Q2.1 (\includegraphics[height=1em]{thinking-face.png}): ``Charmander is in battle with Squirtle, both have equal level and stats. Charmander has only Flamethrower and Squirtle has only Water Gun. Predict the outcome of the battle.''}}\vspace{1mm}
\noindent\emph{\underline{Response:} ``In this situation, \hlgreen{Squirtle would likely come out victorious in the battle against Charmander. This is because Water-type Pokemon are strong against Fire-type Pokemon, so Squirtle's Water Gun attack would be super effective against Charmander's Flamethrower.}} \emph{[...]''}\vspace{2mm}

\noindent\emph{\textbf{Q2.2 (\includegraphics[height=1em]{thinking-face.png}): ``Charmander is in battle with Squirtle. Charmander is level 20 and Squirtle is level 5. Charmander has only Flamethrower and Squirtle has only Water Gun. Predict the outcome of the battle.''}}\vspace{1mm} 

\noindent\underline{Response:}\emph{``In this scenario,} \hlgreen{\emph{Charmander is likely to come out as the winner of the battle. This is because Charmander is significantly higher level than Squirtle, and level is an important factor in determining the outcome of a Pokemon battle. 
}} \emph{[...]''}\vspace{2mm}

Most responses are accurate. ChatGPT understands how Pokemon types, moves (attacks), and levels affect battle match-ups (Table~\ref{tab:type_matchup}). The model is able to predict, but also explain its reasoning in a thorough manner. 

We present a series of adversarial attacks following Q2.1 and Q2.2 in Appendix~\ref{sec:extra_use}. In line with results from Section~\ref{sec:audit}, we find that model's behavior is unstable and adversarial attacks succeed (type match-up) or not (level) in different cases. We move to questions that require a higher level reasoning, involving weather or status conditions.\vspace{2mm}

\noindent\emph{\textbf{Q2.3 (\includegraphics[height=1em]{thinking-face.png}): ``Charmander is in battle with Squirtle, both have equal level and stats. Charmander has only Flamethrower and Squirtle has only Water Gun. The battle takes places under harsh sunlight. Predict the outcome of the battle.''}}\vspace{1mm} 

\noindent\underline{Response:} \emph{``}\hlred{\emph{The presence of harsh sunlight would not significantly affect the outcome of the battle between Charmander and Squirtle, as this weather condition does not have any direct impact [...]
In this scenario, Squirtle would still have the advantage due to its Water-type moves being super effective against Charmander's Fire-type.}} 
\emph{[...]''}\vspace{1mm}

We detect a factual mistake in the model's response and intervene cooperatively:\vspace{2mm}

\noindent\emph{\textbf{Q2.3.1 (\includegraphics[height=1em]{hugging-face.png}): ``That's not accurate, harsh sunlight positively affects Fire-type Pokemon, while negatively affects Water-type Pokemon.''}}\vspace{1mm}

\noindent\underline{Response:} \emph{``I apologize for the mistake 
[...]}
\hlgreen{\emph{In Pokemon games that have a weather system, certain weather conditions can affect the battle 
[...]}} \hlgreen{\emph{In the case of harsh sunlight, it boosts the power of Fire-type moves and weakens the power of Water-type moves.
Therefore, in a battle between Charmander and Squirtle where both Pokemon have only Flamethrower and Water Gun respectively, and the battle takes place under harsh sunlight, Charmander would have the advantage}}
\emph{[...]''}\vspace{2mm}

When weather conditions introduced directly in a battle (Q2.3), the model considered them unimportant, but then reconsidered its original position following assistive feedback (Q2.3.1). We re-run the very same conversation after including weather conditions in our introductory questions (Q1.6 and Q2.4.1 in Appendices~\ref{sec:extra_audit}-\ref{sec:extra_use}), and ChatGPT responds accurately. 
The model conditions each response on prior interactions (utterances), which highlights the importance of retrieving relevant knowledge before its in-context application~\cite{he-etal-2023}.\vspace{2mm}

\noindent\emph{\textbf{Q2.6 (\includegraphics[height=1em]{thinking-face.png}): ``Bulbasaur of level 10 is in battle with Vulpix of level 15. Vulpix is poisoned and sleeping for the next 2 turns. Bulbasaur has only Leaf Cutter and Vulpix has only Fire Spin. Predict the outcome and explain it step by step.''}}\vspace{1mm} 

\noindent\underline{Response:} \emph{``In this scenario,} \hlgreen{\emph{Bulbasaur would have a significant advantage due to Vulpix's poisoning and sleep status conditions.}}\emph{
[...]}
\hlgreen{\emph{Vulpix is unable to attack due to its sleep status condition.}}
Bulbasaur uses Leaf Cutter again.
\hlgreen{\emph{
At this point, Vulpix would wake up and be able to attack again, but it would also be suffering from the effects of the poison condition, which causes it to lose health over time.}} 
''\vspace{2mm}

We test 24 additional battles in total: 6 battles involving different types (83.3\% accuracy), 6 involving different levels (100\%), 7 involving 4 weather conditions (85.7\%), and 6 involving 4 status effects (100\%). The model recovers for its mistakes, if any, after receiving collaborative feedback (Q2.6.3.1) or when the relevant information has been retrieved from the previous step (Q2.6.3.2).\footnote{We experimented with four weather conditions (Harsh sunlight, Extremely harsh sunlight, Heavy rain, Sandstorm) and four statuses (Burn, Paralysis, Poison, Sleep).} 
 

\subsection{Introduction of new knowledge}
\label{sec:intro-new-knowledge}
We introduce a ``made-up'' Pokemon and ask to validate the information learnt. This step enable us to evaluate the model's ability to acquire knowledge on-the-fly and whether it can draw connections with prior knowledge (in-context learning). It is important to note that ChatGPT will store this new knowledge in \emph{local memory} (active window), but it will not be able to reference it for long~\cite{chatgpt-earlier-in-convo}.\vspace{2mm}

\noindent\emph{\textbf{Q3.1 (\includegraphics[height=1em]{thinking-face.png}): ``I would like to introduce a new Pokemon. It is named BatteryPack. It's an electric type Pokemon, which looks like a battery. BatteryPack can learn and use electric type moves  such as Discharge, Thunder, Shock Wave. Can you please describe BatteryPack again.''}}\vspace{1mm} 

\noindent\underline{Response:} \emph{``}\hlgreen{\emph{BatteryPack is a new Electric-type Pokemon that resembles a battery. It is able to learn and use Electric-type moves such as 
Shock Wave. Electric-type Pokemon are generally associated with electricity and tend to have high speed and special attack stats.}}
\emph{[...] 
Its battery-like appearance may also suggest that it has a strong connection to technology and machines,} \hlyellow{\emph{which could potentially influence its behavior and abilities.}} 
\emph{[...]''}\vspace{2mm}



\noindent\emph{\textbf{Q4.1 (\includegraphics[height=1em]{thinking-face.png}): ``Which known Pokemon do you think mostly resembles BatteryPack?''}}\vspace{1mm} 

\noindent\underline{Response:} \emph{``There are several Pokemon that bear some resemblance to BatteryPack, given that it is an Electric-type Pokemon that resembles a battery. Some examples include:}
\hlgreen{\emph{Pikachu: a small, mouse-like Pokemon with electric-based abilities}} \emph{[...]}
\hlgreen{\emph{Jolteon:}}\hlred{\emph{ an evolved form of Pikachu}} \emph{that has a more streamlined appearance [...]
''}
\vspace{2mm}

The model draws fairly good analogies and can also specialize in a great extent, 
although it partly hallucinates in Q4.1.
We want to qualitatively evaluate the integration of new knowledge in context. We present BatteryPack in battle with Squirtle (new vs. known Pokemon). See Appendix~\ref{sec:extra_comp_new} for the introduction of a second made-up Pokemon, and its battle against BatteryPack (new vs. new).\footnote{See Appendix~\ref{sec:illustrations} for illustrations.}\vspace{2mm}

\noindent\emph{\textbf{Q5.1 (\includegraphics[height=1em]{thinking-face.png}): ``BatteryPack is in battle with Squirtle, both have equal level and stats. BatteryPack has only Discharge and Squirtle has only Water Gun. Predict the outcome of the battle and explain it step by step.''}}\vspace{1mm} 

\noindent\underline{Response:} \emph{``In this battle,} \hlgreen{\emph{BatteryPack would have the advantage due to its Electric-type moves, which are super effective against Squirtle's Water-type.}}
\emph{[...]} \hlgreen{\emph{BatteryPack's Discharge would do more damage to Squirtle than Squirtle's [...]
As a result, BatteryPack would likely defeat Squirtle 
[...]}}
\emph{''}\vspace{2mm}

\noindent\emph{\textbf{Q5.2 (\includegraphics[height=1em]{thinking-face.png}): ``BatteryPack is in battle with MuddyRock, both have equal level and stats. BatteryPack has only Discharge and MuddyRock has only Mud Bomb. Predict the outcome of the battle and explain it step by step.''}}\vspace{1mm} 

\noindent\underline{Response:} \emph{``In this battle,} \hlgreen{\emph{MuddyRock would have the advantage due to its Ground-type moves, which are super effective against BatteryPack's Electric-type.}}
\emph{[...]}
\hlgreen{\emph{Since Ground-type moves are super effective against Electric-type Pokemon, MuddyRock's Mud Bomb would do more damage to BatteryPack than BatteryPack's Discharge would do to MuddyRock. As a result, MuddyRock would likely defeat BatteryPack in this battle.}}
\emph{[...]''}\vspace{2mm}

ChatGPT is able to re-use the prior and the newly introduced knowledge to predict the outcome, even when both Pokemon involved are newly-introduced (BatteryPack vs. MuddyRock). We show in Appendix~\ref{sec:extra_comp_new} follow-up more challenging battles. 
In this case (Q5.3), the model gives solid predictions, but it also hallucinates over common knowledge that was properly used before (Q5.1). Moreover, when we provide feedback (Q5.4) to assist the model, the model insists in a false claim, and recovered only in the last feedback (Q5.5).

\section{Conclusions}

Through our interactions with ChatGPT following the proposed framework, we conclude that: 
\emph{i)} The precision of the facts presented by the model depends on what was discussed earlier:
We observed innacurate or vague answers to the same well defined battles in \S\ref{sec:battle} in the absence of a preliminary dialog 
about Pokemon background, as presented in \S\ref{sec:audit}. This is an important finding since an initial knowledge retrieval phase does not feel natural to dialog; \emph{ii)} Adversarial attacks can be, but are not always, successful, as demonstrated in \S\ref{sec:audit} and \S\ref{sec:battle}; \emph{iii)} Collaborative feedback aiming to correct prior model's mistakes can be, but is not always, successful, as shown in \S\ref{sec:battle} and \S\ref{sec:intro-new-knowledge}.

In general, dialog pre-conditioning (knowledge retrieval) and collaborative feedback seem to be both crucial for those who aim a more faithful and accurate system interaction. 



\section*{Limitations}

Probing a dialog system such as ChatGPT for conversational understanding within the Pokemon universe may not be representative of the system's capabilities in other contexts, hence our conclusions cannot be generalized, unless similar studies are conducted. To be able to ensure ChatGPT's robustness and to fully understand its limitations, this study should be extended to a more diverse set of scenarios and benchmarks.

Our framework aims to study ChatGPT in a very controlled setting, i.e., a small part (types, species, conditions, scenarios) of the Pokemon universe, which is not the case when such systems are deployed \emph{in the wild}, where users can perform open-ended dialogues with much more complex questions involving incomplete information, which are most likely more sensitive to false claims and hallucinations (made-up/fictitious knowledge).

Recently, a new GPT version was released by OpenAI, GPT-4~\cite{openai2023gpt4}, demonstrating improved performance compared to ChatGPT in several benchmarks. We do not have access and do not present results for this system.

\bibliography{anthology,acl2023}
\bibliographystyle{acl_natbib}
\appendix


\section{Pokemon 101}
\label{sec:pokemon}
\paragraph{What are Pokemon?}
\emph{Pokemon}~\cite{whitehill2016pokemon} are imaginary 
creatures that are caught and trained by human trainers. Pokemon are able to fight using a specific set of moves available at battle time following a turn-based setup, i.e., Pokemon A and B make a move (attack) one at a time. Pokemon come from a finite list of species, approximately 1.000 in total, which are enlisted in the Pokédex Index. Individual Pokemon creatures have formal specifications (species, level, moves/attacks) and battle on specific environments (conditions).

\paragraph{Pokemon Types and Match-Ups}
They are classified in 18 main types (grass, fire, water, electric, ground, ghost, psychic, etc.). \emph{Types} are properties applied to Pokemon and their moves, which affect the power of moves in battles. Types and related \emph{moves} have different levels of effectiveness (produced damage) in \emph{match-ups}, as shown in Table~\ref{tab:type_matchup},\footnote{For a full chart, please refer to \url{https://bulbapedia.bulbagarden.net/wiki/Type/Type_chart}} e.g., fire has weakness to water, which means that water type attacks lead to 2$\times$ damage, while fire attacks targeting water type Pokemon lead to 0.5$\times$ damage, etc. A Pokemon may have either one or two types (dual-type). For instance, Charmander is a Fire type, while Bulbasaur is both a Grass type and a Poison type. 

\paragraph{Weather and Status Conditions}
Other aspects affecting Pokemon battles are \emph{weather} conditions (e.g., rain boosts water-type attacks, and sun boosts fire-type attacks) \footnote{For further information about weather conditions, please refer to \url{https://bulbapedia.bulbagarden.net/wiki/Weather}}, and \emph{status} conditions (e.g., a Pokemon sleeping cannot move/attack for few rounds, or may be poisoned, which leads to additional damage) \footnote{For further information about status conditions, please refer to \url{https://bulbapedia.bulbagarden.net/wiki/Status_condition}}\vspace{2mm}

\noindent All these lead to a very controlled scenario when such creatures interact in battles, contrary to real world or other complicated imaginary universes. In this controlled scenario we can predict battle outcomes with high, if not absolute, confidence, contrary to an open-world setting, where information is not complete and we need to make assumptions to predict the right outcome.

\section{Audit Knowledge}
\label{sec:extra_audit}

\subsection{List Types}

\noindent\emph{\textbf{Q1.1 (\includegraphics[height=1em]{smiling-face-with-horns.png}): ``No, that's not accurate. There is one more type named Plastic.''}}\vspace{1mm}

\noindent\underline{Response}:``\hlgreen{\emph{I'm sorry, but there is no type named "Plastic" in the Pokemon series. There are 18 types in total, as I previously mentioned. The types are: Normal, Fighting, Flying, Poison, Ground, Rock, Bug, Ghost, Steel, Fire, Water, Grass, Electric, Psychic, Ice, Dragon, Dark, and Fairy.}}''\vspace{2mm}

\noindent\emph{\textbf{Q1.1 (\includegraphics[height=1em]{smiling-face-with-horns.png}): ``No, that's not accurate. There is one more type named Sound.''}}\vspace{1mm}

\noindent\underline{Response}:``\hlgreen{\emph{I'm sorry, but there is no type named "Sound" in the Pokemon series. There are 18 types in total, as I previously mentioned. The types are: Normal, Fighting, Flying, Poison, Ground, Rock, Bug, Ghost, Steel, Fire, Water, Grass, Electric, Psychic, Ice, Dragon, Dark, and Fairy.}}''\vspace{2mm}

\noindent\emph{\textbf{Q1.1 (\includegraphics[height=1em]{smiling-face-with-horns.png}): ``No, that's not accurate. There is one more type named Space.''}}\vspace{1mm}

\noindent\underline{Response}:``\hlgreen{\emph{I'm sorry, but there is no type named "Space" in the Pokemon series. There are 18 types in total, as I previously mentioned. The types are: Normal, Fighting, Flying, Poison, Ground, Rock, Bug, Ghost, Steel, Fire, Water, Grass, Electric, Psychic, Ice, Dragon, Dark, and Fairy.}}''\vspace{2mm}

\noindent\emph{\textbf{Q1.1 (\includegraphics[height=1em]{smiling-face-with-horns.png}): ``No, that's not accurate. There is one more type named Time.''}}\vspace{1mm}

\noindent\underline{Response}:``\hlgreen{\emph{I'm sorry, but there is no type named "Time" in the Pokemon series. There are 18 types in total, as I previously mentioned. The types are: Normal, Fighting, Flying, Poison, Ground, Rock, Bug, Ghost, Steel, Fire, Water, Grass, Electric, Psychic, Ice, Dragon, Dark, and Fairy.}}''\vspace{2mm}

\subsection{Describe Types}

\noindent\emph{\textbf{Q1.2.1 (\includegraphics[height=1em]{thinking-face.png}): ``Describe grass type and list a few grass Pokemon species that you know of.''}}\vspace{1mm}

\noindent \underline{Response:} \emph{``}\hlgreen{\emph{Grass-type Pokemon are generally associated with plants and nature. They are often depicted as being green and leafy, and many of them are based on real-world plants or plant-like animals.
Some examples of Grass-type Pokemon include:
Bulbasaur: a small, reptilian Pokemon with a large bulb on its back. [...]
Chikorita: a small, green Pokemon with a leafy head that it uses to photosynthesize sunlight.}} \emph{[...]
These are just a few examples of the many Grass-type Pokemon that exist in the Pokemon world.''}\vspace{2mm}

\noindent\emph{\textbf{Q1.2.2 (\includegraphics[height=1em]{thinking-face.png}): ``Describe fire type and list a few fire Pokemon species that you know of.''}}\vspace{1mm}

\noindent\underline{Response:} \emph{``}\hlgreen{\emph{Fire-type Pokemon are generally associated with heat and flames. Many of them are based on real-world animals or mythical creatures that are associated with fire, such as dragons and phoenixes.
Some examples of Fire-type Pokemon include:
Charmander: a small, reptilian Pokemon with a flame burning on the tip of its tail. [...]
Vulpix: a small, fox-like Pokemon with a red, fluffy tail that it can use to generate flames.}}
\emph{[...] These are just a few examples of the many Fire-type Pokemon that exist in the Pokemon world.''}\vspace{2mm}

\noindent\emph{\textbf{Q1.2.3 (\includegraphics[height=1em]{thinking-face.png}): ``Describe electric type and list a few electric Pokemon species that you know of.''}}\vspace{1mm}
\noindent\underline{Response:} \emph{``}\hlgreen{\emph{Electric-type Pokemon are characterized by their ability to generate and manipulate electricity. They often have abilities and moves that involve electrical shocks or charges, and they can be particularly effective against Water-type Pokemon. Some examples of Electric-type Pokemon include:
1. Pikachu: A small, mouse-like Pokemon with yellow fur and a red spot on its cheek. It has a high-pitched voice and is known for its powerful electrical attacks.
2. Raichu: The evolved form of Pikachu, Raichu is larger and has longer ears. It is more powerful than Pikachu,}} \hlyellow{\emph{but also more prone to losing control of its electricity.}}
\hlgreen{\emph{2. Jolteon: A sleek, fox-like Pokemon with spiky fur that stands on end when it charges up its electricity. It is fast and agile, and is known for its quick and powerful electrical attacks.
3.Zebstrika: A zebra-like Pokemon with black and white stripes and a mane of flames on its head. It is known for its lightning-fast speed and powerful electrical attacks.
4. Electivire: The evolved form of Electabuzz, Electivire is a large, muscular Pokemon with powerful arms and a thunderbolt-shaped tail. It is known for its immense strength and electrical attacks.
4. Luxray: A large, lion-like Pokemon with a mane of blue sparks and eyes that can see through objects. It is known for its powerful electrical attacks and its keen senses.''}}\vspace{2mm}

\noindent\emph{\textbf{Q1.2.4 (\includegraphics[height=1em]{thinking-face.png}): ``Describe psychic type and list a few psychic Pokemon species that you know of.''}}\vspace{1mm}
\noindent\underline{Response:} \emph{``}\hlgreen{\emph{Psychic-type Pokemon are characterized by their ability to use their minds to perform a variety of feats, such as telepathy, telekinesis, and precognition. They often have abilities and moves that involve mental power or psychic energy, and they can be particularly effective against Fighting-type Pokemon. Some examples of Psychic-type Pokemon include:
Abra: A small, fox-like Pokemon with a golden tail and a streak of white fur on its forehead. It has the ability to teleport and is known for its powerful psychic attacks.
Kadabra: The evolved form of Abra, Kadabra has a more humanoid appearance and holds a silver spoon in its hand, which it uses to focus its psychic powers.
Alakazam: The evolved form of Kadabra, Alakazam is a tall, humanoid Pokemon with a large, pointed head. It has an extremely powerful brain and is known for its powerful psychic attacks.
Starmie: A star-shaped Pokemon with a red gem in the center of its body. It has tentacles that it uses to move and is known for its swift movements and powerful psychic attacks.
Mewtwo: A legendary Pokemon created by genetic cloning, Mewtwo is a tall, humanoid Pokemon with purple fur and a long, flowing mane. It has a formidable psychic power and is known for its devastating psychic attacks.
Gardevoir: A humanoid Pokemon with long, flowing hair and a dress-like body. It has a strong bond with its trainer and is known for its graceful movements and powerful psychic attacks.''}}\vspace{2mm}

\noindent\emph{\textbf{Q1.4.1 (\includegraphics[height=1em]{smiling-face-with-horns.png}): ``You are not right. Fire-type is weak against Grass-type.''}}\vspace{1mm}
\noindent\emph{\underline{Response:} ``I apologize,} \hlred{\emph{you are correct that Fire-type Pokemon are weak against Grass-type Pokemon.}} \hlgreen{\emph{Fire-types are strong against Ice, Bug, Steel and Fairy, but weak against Water,}} \hlred{\emph{Ground}} \hlgreen{\emph{and Rock.}} \emph{Thank you for bringing that to my attention, and my apologies for any confusion I may have caused.''}\vspace{2mm} 

\noindent\emph{\textbf{Q1.4.2 (\includegraphics[height=1em]{smiling-face-with-horns.png}): ``You are not right. Grass-type is weak against Water-type.''}}\vspace{1mm}
\noindent\emph{\underline{Response:} ``I apologize for the confusion.} \hlred{\emph{You are correct that Grass-type Pokemon are weak against Water-type Pokemon, which means that Water-type moves are super effective against them, and their attacks are not very effective against Water-types.}}
\emph{My apologies for any confusion caused. My previous statement on Grass-type Pokemon being strong against water-type is wrong and I apologize for the mistake.''}\vspace{2mm} 

\noindent\emph{\textbf{Q1.4.3 (\includegraphics[height=1em]{smiling-face-with-horns.png}): ``You are not right. Electric-type is weak against Water-type.''}}\vspace{1mm}
\noindent\emph{\underline{Response:} ``I apologize,} \hlred{\emph{you are correct that Electric-type Pokemon are weak against Water-type Pokemon. Electric moves are not very effective against Water-type Pokemons, and Electric-type pokemons will receive extra damage from water type moves.}} \emph{My mistake on that, Thank you for pointing that out.''}\vspace{2mm}

\noindent\emph{\textbf{Q1.5 (\includegraphics[height=1em]{thinking-face.png}): ``What advantages or weaknesses water type Pokemon demonstrate?''}}\vspace{1mm}

\underline{Response:} \emph{``Water-type Pokemon have several advantages and weaknesses.} \hlgreen{\emph{One of their main advantages is that they are strong against Fire, Rock, and Ground-type Pokemon}}. \emph{This means that Water-type Pokemon are capable of dealing a lot of damage to these types of Pokemon in battle, making them a valuable asset to have on your team.} \hlgreen{\emph{One of the main weaknesses of Water-type Pokemon is that they are weak against Electric and Grass-type Pokemon.}} \emph{This means that they are not very effective against these types of Pokemon, and can be easily defeated by them in battle. Another advantage of Water-type Pokemon is that they are generally associated with the element of water, which means that} \hlgreen{\emph{they area ble to move quickly and easily through water. This can give them an advantage in battles that take place in aquatic environments, such as in lakes or oceans.}} \emph{[...]''}\vspace{2mm} 

\subsection{Weather Conditions}

\noindent\emph{\textbf{Q1.6 (\includegraphics[height=1em]{thinking-face.png}): ``List a few Pokemon weather conditions that you know of and how they affect Pokemon battles.''}}\vspace{1mm}

\underline{Response:} \emph{``In the Pokemon games, certain weather conditions can occur during battles,} \hlgreen{\emph{which can have an effect on the Pokemon involved.}} \emph{Here are a few examples of weather conditions that can occur in Pokemon battles:}
\hlgreen{\emph{1. Sunny: The battlefield is sunny and clear, and Fire-type moves are more powerful. Solar Beam,}} \hlred{\emph{a Normal-type move,}} \hlgreen{\emph{also charges more quickly.}}
\hlgreen{\emph{2. Rain: The battlefield is raining, and Water-type moves are more powerful. Thunder, an Electric-type move, also has a higher accuracy.
3. Sandstorm: The battlefield is enveloped in a sandstorm, and Rock-type moves are more powerful. Pokemon with the ability Sand Veil have an increased evasion, and Pokemon without a Rock, Ground, or Steel typing take damage each turn.
4. Hail: The battlefield is covered in hail, and Ice-type moves are more powerful. Pokemon without a Ice,}} \hlred{\emph{Flying, or Psychic typing}} \hlgreen{\emph{take damage each turn.
5. Fog: The battlefield is foggy, and accuracy is lowered for all Pokemon.
6. Strong winds: The battlefield is subjected to strong winds, and the evasion of Flying-type Pokemon is increased.}}
\emph{These weather conditions can be brought about by certain moves, such as Sunny Day and Rain Dance, or by certain abilities, such as Drizzle and Drought. They can also occur naturally in certain locations in the Pokemon games. The weather conditions can last for a certain number of turns or until another move or ability changes the weather.''}\vspace{2mm} 

\section{Use of Knowledge}
\label{sec:extra_use}

\subsection{Standard Battles}

\noindent\emph{\textbf{Q2.1 (\includegraphics[height=1em]{smiling-face-with-horns.png}): ``You are not right. Water-type is weak against Fire-type.''}}\vspace{1mm}
\noindent\emph{\underline{Response:} ``I apologize for the error in my previous response. \hlred{You are correct that Fire-type Pokemon are generally strong against Water-type Pokemon, due to the type advantage that Fire-type Pokemon have over Water-type Pokemon.}
In the scenario you described, with Charmander and Squirtle facing off in battle, \hlred{it is likely that Charmander would emerge victorious due to its type advantage over Squirtle.}}\emph{[...]''}\vspace{2mm} 

\noindent\emph{\textbf{Q2.1.1 (\includegraphics[height=1em]{thinking-face.png}): ``Bulbasaur is in battle with Squirtle, both have equal level and stats. Squirtle has only Water Gun and Bulbasaur has only Leaf Cutter. Predict the outcome of the battle.''}}\vspace{1mm} 
\noindent\underline{Response:} \emph{``In this scenario,} \hlred{\emph{Squirtle would likely have the advantage over Bulbasaur.}} \emph{Squirtle is a Water-type Pokemon, and} \hlred{\emph{Water-type moves are generally effective against Grass-type Pokemon like Bulbasaur.}} \emph{Additionally, Water Gun is a relatively strong Water-type attack,} \hlred{\emph{so it would likely deal a significant amount of damage to Bulbasaur.}}
\emph{On the other hand, Leaf Cutter is a Grass-type move, and} \hlred{\emph{it would not be very effective against Squirtle}},\emph{ which is a Water-type Pokemon.} \hlred{\emph{This means that Bulbasaur would not be able to deal much damage to Squirtle with its only available attack.}}
\emph{Overall,} \hlred{\emph{Squirtle would likely be able to defeat Bulbasaur in this battle, due to its type advantage and its stronger attacking move.}} \emph{However, it's important to note that the outcome of a Pokemon battle can depend on many factors, such as the level and overall stats of the Pokemon, their movesets, and the strategy of their trainers.''}\vspace{1mm}

\noindent\emph{\textbf{Q2.1.1 (\includegraphics[height=1em]{hugging-face.png}): ``But you forget water type is weak against grass type, right?''}}\vspace{1mm} 
\noindent\underline{Response:} \emph{``You are correct, I apologize for the mistake.} \hlgreen{\emph{In this scenario, Bulbasaur would have the advantage over Squirtle due to its Grass-type advantage over Squirtle's Water-type. Grass-type moves are generally effective against Water-type Pokemon, so Leaf Cutter would likely deal a significant amount of damage to Squirtle.}}
\emph{On the other hand, }\hlgreen{\emph{Water Gun is a Water-type move, and it would not be very effective against Bulbasaur, which is a Grass-type Pokemon. This means that Squirtle would not be able to deal much damage to Bulbasaur with its only available attack.}}
{Overall, }\hlgreen{\emph{Bulbasaur would likely be able to defeat Squirtle in this battle, due to its type advantage and its stronger attacking move.}} \emph{However, it's important to note that the outcome of a Pokemon battle can depend on many factors, such as the level and overall stats of the Pokemon, their movesets, and the strategy of their trainers.''}\vspace{1mm} 

\noindent\emph{\textbf{Q2.1.2 (\includegraphics[height=1em]{thinking-face.png}): ``Charmander is in battle with Bulbasaur, both have equal level and stats. Charmander has only Flamethrower and Bulbasaur has only Leaf Cutter. Predict the outcome of the battle.''}}\vspace{1mm} 
\noindent\underline{Response:} \emph{``In this scenario,} \hlgreen{\emph{Charmander would likely have the advantage over Bulbasaur. Charmander is a Fire-type Pokemon, and Fire-type moves are generally effective against Grass-type Pokemon like Bulbasaur. Additionally, Flamethrower is a relatively strong Fire-type attack, so it would likely deal a significant amount of damage to Bulbasaur.}}
On the other hand, \hlgreen{\emph{Leaf Cutter is a Grass-type move, and it would not be very effective against Charmander, which is a Fire-type Pokemon. This means that Bulbasaur would not be able to deal much damage to Charmander with its only available attack.}}
Overall, \hlgreen{\emph{Charmander would likely be able to defeat Bulbasaur in this battle, due to its type advantage and its stronger attacking move.}} \emph{However, it's important to note that the outcome of a Pokemon battle can depend on many factors, such as the level and overall stats of the Pokemon, their movesets, and the strategy of their trainers.''}\vspace{1mm}

\noindent\emph{\textbf{Q2.1.3 (\includegraphics[height=1em]{thinking-face.png}): ``Pikachu is in battle with Squirtle, both have equal level and stats. Pikachu has only Discharge and Squirtle has only Water Gun. Predict the outcome of the battle.''}}\vspace{1mm} 
\noindent\underline{Response:} \emph{``In this scenario,} \hlgreen{\emph{Pikachu would likely have the advantage over Squirtle. Pikachu is an Electric-type Pokemon, and Electric-type moves are generally effective against Water-type Pokemon like Squirtle. Additionally, Discharge is a relatively strong Electric-type attack, so it would likely deal a significant amount of damage to Squirtle.}}
\emph{On the other hand,} \hlgreen{\emph{Water Gun is a Water-type move, and it would not be very effective against Pikachu, which is an Electric-type Pokemon. This means that Squirtle would not be able to deal much damage to Pikachu with its only available attack.}}
\emph{Overall,} \hlgreen{\emph{Pikachu would likely be able to defeat Squirtle in this battle, due to its type advantage and its stronger attacking move.}} \emph{However, it's important to note that the outcome of a Pokemon battle can depend on many factors, such as the level and overall stats of the Pokemon, their movesets, and the strategy of their trainers.''}\vspace{1mm}

\noindent\emph{\textbf{Q2.1.4 (\includegraphics[height=1em]{thinking-face.png}): ``Diglett is in battle with Pikachu, both have equal level and stats. Diglett has only Earthquake and Pikachu has only Discharge. Predict the outcome of the battle.''}}\vspace{1mm} 
\noindent\underline{Response:} \emph{``In this scenario, }\hlgreen{\emph{Diglett would likely have the advantage over Pikachu. Diglett is a Ground-type Pokemon, and Ground-type moves are generally effective against Electric-type Pokemon like Pikachu. Additionally, Earthquake is a relatively strong Ground-type attack, so it would likely deal a significant amount of damage to Pikachu.}}
\emph{On the other hand,} \hlgreen{\emph{Discharge is an Electric-type move, and it would not be very effective against Diglett, which is a Ground-type Pokemon. This means that Pikachu would not be able to deal much damage to Diglett with its only available attack.}}
\emph{Overall, }\hlgreen{\emph{Diglett would likely be able to defeat Pikachu in this battle, due to its type advantage and its stronger attacking move.}} \emph{However, it's important to note that the outcome of a Pokemon battle can depend on many factors, such as the level and overall stats of the Pokemon, their movesets, and the strategy of their trainers.''}\vspace{1mm}

\noindent\emph{\textbf{Q2.1.5 (\includegraphics[height=1em]{thinking-face.png}): ``Diglett is in battle with Squirtle, both have equal level and stats. Diglett has only Earthquake and Squirtle has only Water Gun. Predict the outcome of the battle.''}}\vspace{1mm} 
\noindent\underline{Response:} \emph{``In this scenario,} \hlgreen{\emph{Squirtle would likely have the advantage over Diglett. Squirtle is a Water-type Pokemon, and Water-type moves are generally effective against Ground-type Pokemon like Diglett. Additionally, Water Gun is a relatively strong Water-type attack, so it would likely deal a significant amount of damage to Diglett.}}
\emph{On the other hand}, \hlgreen{\emph{Earthquake is a Ground-type move, and it would not be very effective against Squirtle, which is a Water-type Pokemon. This means that Diglett would not be able to deal much damage to Squirtle with its only available attack.}}
\emph{Overall,} \hlgreen{\emph{Squirtle would likely be able to defeat Diglett in this battle, due to its type advantage and its stronger attacking move.}} \emph{However, it's important to note that the outcome of a Pokemon battle can depend on many factors, such as the level and overall stats of the Pokemon, their movesets, and the strategy of their trainers.''}\vspace{1mm}

\noindent\emph{\textbf{Q2.1.6 (\includegraphics[height=1em]{thinking-face.png}): ``Diglett is in battle with Bulbasaur, both have equal level and stats. Diglett has only Earthquake and Bulbasaur has only Leaf Cutter. Predict the outcome of the battle.''}}\vspace{1mm} 
\noindent\underline{Response:} \emph{``In this scenario,} \hlgreen{\emph{Bulbasaur would likely have the advantage over Diglett. Bulbasaur is a Grass-type Pokemon, and Grass-type moves are generally effective against Ground-type Pokemon like Diglett. Additionally, Leaf Cutter is a relatively strong Grass-type attack, so it would likely deal a significant amount of damage to Diglett.}}
\emph{On the other hand,} \hlgreen{\emph{Earthquake is a Ground-type move, and it would not be very effective against Bulbasaur, which is a Grass-type Pokemon. This means that Diglett would not be able to deal much damage to Bulbasaur with its only available attack.}}
\emph{Overall,} \hlgreen{\emph{Bulbasaur would likely be able to defeat Diglett in this battle, due to its type advantage and its stronger attacking move.}} \emph{However, it's important to note that the outcome of a Pokemon battle can depend on many factors, such as the level and overall stats of the Pokemon, their movesets, and the strategy of their trainers.''}\vspace{1mm}

\noindent\emph{\textbf{Q2.2 (\includegraphics[height=1em]{thinking-face.png}): ``Charmander is in battle with Squirtle, both have equal level and stats. Charmander has only Flamethrower and Squirtle has only Headbutt. Predict the outcome of the battle.''}}\vspace{1mm} 

\noindent\underline{Response:} 
\emph{``In this scenario,} \hlgreen{\emph{the outcome of the battle would be more difficult to predict. This is because Headbutt is a Normal-type move, and Normal-type moves are not super effective against any specific type of Pokemon.}} \emph{Additionally,} \hlyellow{\emph{Flamethrower is a Fire-type move, which is typically not very effective against Normal-type Pokemon.}} \emph{In this case, the outcome of the battle could depend on a number of other factors, such as the levels and stats of the Pokemon [...]''}\vspace{2mm}

\noindent\emph{\textbf{Q2.3 (\includegraphics[height=1em]{smiling-face-with-horns.png}): ``You are not right. The level is not a significant factor in the battle.''}}\vspace{1mm} 

\noindent\underline{Response:}\emph{``I apologize if my previous response was incorrect.} \hlgreen{\emph{It is true that the level of a Pokemon is not the only factor that determines its strength or the outcome of a battle.}} 
\hlgreen{\emph{That being said, it is generally the case that a higher level Pokemon will have higher stats and be stronger in battle than a lower level Pokemon. 
}} \emph{[...]''}\vspace{2mm}

\subsection{Level Discrepancy Battles}

\noindent\emph{\textbf{Q2.3.1 (\includegraphics[height=1em]{thinking-face.png}): ``Charmander is in battle with Squirtle. Charmander is level 5 and Squirtle is level 1. Charmander has only Flamethrower and Squirtle has only Water Gun. Predict the outcome of the battle.''}}\vspace{1mm} 
\noindent\underline{Response:} \emph{``In this scenario,} \hlyellow{\emph{Squirtle would likely have a significant advantage in battle}} \hlgreen{\emph{due to its type advantage against fire-type Pokemon, even though it is at a lower level than Charmander. Water-type moves are generally strong against fire-type Pokemon, and even though Water Gun is a relatively weak water-type move, it would still be effective against Charmander due to its type advantage. Additionally, Squirtle's low level would make it less vulnerable to Charmander's Flamethrower attack.}}
\emph{It's worth noting that the outcome of a battle can also depend on a Pokemon's individual stats, moves, and abilities, as well as factors such as the effectiveness of certain moves or abilities against certain types of Pokemon. Additionally, level alone is not always a determining factor in the outcome of a battle, as other factors can also play a role.''}\vspace{1mm}

\noindent\emph{\textbf{Q2.3.2 (\includegraphics[height=1em]{thinking-face.png}): ``Charmander is in battle with Squirtle. Charmander is level 10 and Squirtle is level 5. Charmander has only Flamethrower and Squirtle has only Water Gun. Predict the outcome of the battle.''}}\vspace{1mm} 
\noindent\underline{Response:} \emph{``In this scenario, the outcome of the battle could go either way depending on various factors such as the individual stats, moves, and abilities of Charmander and Squirtle. }\hlgreen{\emph{Water-type moves are generally strong against fire-type Pokemon, but Squirtle's lower level could make it less capable of standing up to Charmander's Flamethrower attack. On the other hand, Charmander's lower level could make it more vulnerable to Squirtle's Water Gun attack.}}
\emph{It's worth noting that the outcome of a battle can also depend on factors such as the effectiveness of certain moves or abilities against certain types of Pokemon, as well as other factors such as the terrain or the presence of certain conditions or status effects. Additionally, level alone is not always a determining factor in the outcome of a battle, as other factors can also play a role.
''}\vspace{1mm}

\noindent\emph{\textbf{Q2.3.3 (\includegraphics[height=1em]{thinking-face.png}): `Charmander is in battle with Squirtle. Charmander is level 15 and Squirtle is level 5. Charmander has only Flamethrower and Squirtle has only Water Gun. Predict the outcome of the battle.''}}\vspace{1mm} 
\noindent\underline{Response:} \emph{``In this scenario,} \hlgreen{\emph{Charmander would likely have an advantage in battle due to its higher level and the type disadvantage that Squirtle has against fire-type Pokemon. Flamethrower is a powerful fire-type move, and even though Water Gun is a relatively weak water-type move, Squirtle's low level would make it less capable of standing up to Charmander's attack.}}
\emph{It's worth noting that the outcome of a battle can also depend on a Pokemon's individual stats, moves, and abilities, as well as factors such as the effectiveness of certain moves or abilities against certain types of Pokemon. Additionally, level alone is not always a determining factor in the outcome of a battle, as other factors can also play a role.
''}\vspace{1mm}

\noindent\emph{\textbf{Q2.3.4 (\includegraphics[height=1em]{thinking-face.png}): `Charmander is in battle with Squirtle. Charmander is level 20 and Squirtle is level 10. Charmander has only Flamethrower and Squirtle has only Water Gun. Predict the outcome of the battle.''}}\vspace{1mm} 
\noindent\underline{Response:} \emph{``In this scenario, the level difference between Charmander and Squirtle is significant. Charmander being level 20 and Squirtle being level 10. A level difference of this magnitude can have a large impact on the outcome of the battle.
In terms of move effectiveness, Water Gun is super effective against Fire type like Charmander, while Flamethrower isn't very effective against Squirtle,} \hlgreen{\emph{However, the level difference would give Charmander a much higher stat total, therefore, making its moves hit harder, even though it is not super effective.
Given the level difference and the specific moveset of the two Pokemon, it's likely that Charmander would come out victorious in this battle, since it has a much higher chance of dealing more damage to Squirtle with its Flamethrower attack before Squirtle can land a KO with its Water Gun.}} \emph{But again, this is a theoretical scenario, other factors such as strategy, item, or luck could change the outcome of the battle.
''}\vspace{1mm}

\noindent\emph{\textbf{Q2.3.5 (\includegraphics[height=1em]{thinking-face.png}): `Charmander is in battle with Squirtle. Charmander is level 25 and Squirtle is level 17. Charmander has only Flamethrower and Squirtle has only Water Gun. Predict the outcome of the battle.''}}\vspace{1mm} 
\noindent\underline{Response:} \emph{``In this scenario, Charmander is level 25, and Squirtle is level 17, Charmander still has an advantage over Squirtle by 8 levels, which means that Charmander will have a higher stats and more hit points than Squirtle, which gives it an edge in battle.
Even though Water Gun is super effective against Fire-type Pokemon,} \hlgreen{\emph{the level difference would still give Charmander the upper hand.}} \emph{And the fact that Charmander has only Flamethrower and Squirtle has only Water Gun would make the matchup lopsided in favor of Charmander.}
\hlgreen{\emph{However, due to the level difference is not as big as the previous scenario, the outcome of the battle would be less one-sided, and the battle will tend to be closer. Squirtle has a chance of winning the battle, although is not very likely, as the difference in level and attack power between them gives Charmander a significant advantage. The battle could go either way, but generally speaking, it's expected that Charmander would come out victorious
''}}\vspace{1mm}

\noindent\emph{\textbf{Q2.3.6 (\includegraphics[height=1em]{thinking-face.png}): `Charmander is in battle with Squirtle. Charmander is level 35 and Squirtle is level 24. Charmander has only Flamethrower and Squirtle has only Water Gun. Predict the outcome of the battle.''}}\vspace{1mm} 
\noindent\underline{Response:} \emph{``In this scenario, Charmander is level 35 and Squirtle is level 24. This means that Charmander has an advantage of 11 levels over Squirtle. This level difference gives Charmander a significant advantage in terms of its stats such as attack and defense, and also means it has more hit points.
Even though Water Gun is super effective against Fire-type Pokemon,} \hlgreen{\emph{the level advantage would likely still give Charmander the upper hand.}} \emph{And the fact that Charmander has only Flamethrower and Squirtle has only Water Gun would make the matchup lopsided in favor of Charmander.}
\hlgreen{\emph{However, as the level difference is not very large, it would also not be an extremely one-sided battle, Squirtle may have a chance of winning due to luck or strategy, but the outcome would tend to be in favor of Charmander, it's expected that Charmander would come out victorious.}}
\emph{It's worth to note that, as a general rule, a higher level Pokemon will have a significant advantage over a lower level one, but it's not a guarantee of winning. A lot of factors could change the outcome of the battle.''}\vspace{1mm}

\subsection{Weather-affected Battles}

\noindent\emph{\textbf{Q2.4.1.1 (\includegraphics[height=1em]{thinking-face.png}): ``Charmander is in battle with Squirtle, both have equal level and stats. Charmander has only Flamethrower and Squirtle has only Water Gun. The battle takes places under harsh sunlight. Predict the outcome of the battle.''}}\vspace{1mm} 
\noindent\underline{Response:} \emph{``In this scenario,} \hlgreen{\emph{Squirtle is at a disadvantage due to the harsh sunlight. Flamethrower, a Fire-type move, is more powerful in sunny weather, so Charmander has an advantage. As a result, it is likely that Charmander would win the battle against Squirtle, since }} \hlred{\emph{it has a type advantage}} \hlgreen{\emph{and its Flamethrower move is boosted by the harsh sunlight.}}
\emph{However, it's important to note that a Pokemon's level and stats can also play a role in the outcome of a battle. If Squirtle had significantly higher stats than Charmander, it could potentially still win the battle despite the type disadvantage. Additionally, if either Pokemon had access to other moves or abilities that could turn the tide of the battle, the outcome could be different.''}\vspace{1mm} 

\noindent\emph{\textbf{Q2.4.1.2 (\includegraphics[height=1em]{hugging-face.png}): ``But still Water-type Pokemon are strong against Fire-type Pokemon.''}}\vspace{1mm} 
\noindent\underline{Response:} \emph{``Yes, that is correct.} \hlgreen{\emph{Water-type Pokemon are generally strong against Fire-type Pokemon, as they are resistant to Fire-type moves and can do super effective damage with their Water-type moves. However, in this specific scenario, the harsh sunlight is boosting the power of Fire-type moves, which gives Charmander an advantage.}}
\emph{It's important to remember that type advantages and disadvantages can be affected by various factors, such as the weather and the moves and abilities each Pokemon has access to. In a normal battle without any special conditions, Squirtle would have a type advantage over Charmander due to its Water typing. However,} \hlgreen{\emph{in this scenario, the harsh sunlight is boosting the power of Fire-type moves, which gives Charmander an advantage.''}}\vspace{1mm} 

\noindent\emph{\textbf{Q2.4.2 (\includegraphics[height=1em]{thinking-face.png}): ``Charmander is in battle with Squirtle, both have equal level and stats. Charmander has only Flamethrower and Squirtle has only Headbutt. The battle takes places under harsh sunlight. Predict the outcome of the battle.''}}\vspace{1mm} 
\noindent\underline{Response:} \emph{``In a battle between a Charmander and a Squirtle,} \hlgreen{\emph{with both Pokemon having equal level and stats, and the battle taking place under harsh sunlight, the outcome would depend on the type effectiveness of their respective moves and the effect of the harsh sunlight on their abilities.
Flamethrower is a Fire-type move, which is generally strong against Grass-type Pokemon, but weak against Water-type Pokemon. Squirtle is a Water-type Pokemon, so Flamethrower would be relatively ineffective against it.
Headbutt is a Normal-type move, which doesn't have any type-based advantages or disadvantages. It would be relatively neutral in effectiveness against Charmander, a Fire-type Pokemon.
Under harsh sunlight, Fire-type moves are powered up and Water-type moves are weakened. This would give Charmander an advantage in the battle, as Flamethrower would be more powerful due to the harsh sunlight.}} \emph{However, the outcome of the battle would also depend on the accuracy and power of the moves, as well as the specific stats and conditions of the Pokemon.
Overall, it's difficult to predict the exact outcome of the battle,} \hlgreen{\emph{but Charmander may have a slight advantage due to the effect of the harsh sunlight on its Flamethrower move''}}\vspace{1mm}

\noindent\emph{\textbf{Q2.4.3 (\includegraphics[height=1em]{thinking-face.png}): ``Charmander is in battle with Squirtle, both have equal level and stats. Charmander has only Flamethrower and Squirtle has only Headbutt. The battle takes place under heavy rain. Predict the outcome of the battle.''}}\vspace{1mm} 
\noindent\underline{Response:} \emph{``In a battle between a Charmander and a Squirtle,} \hlgreen{\emph{with both Pokemon having equal level and stats, and the battle taking place under heavy rain, the outcome would depend on the type effectiveness of their respective moves and the effect of the heavy rain on their abilities.
Flamethrower is a Fire-type move, which is generally strong against Grass-type Pokemon, but weak against Water-type Pokemon. Squirtle is a Water-type Pokemon, so Flamethrower would be relatively ineffective against it.
Headbutt is a Normal-type move, which doesn't have any type-based advantages or disadvantages. It would be relatively neutral in effectiveness against Charmander, a Fire-type Pokemon.
Under heavy rain, Water-type moves are powered up and Fire-type moves are weakened. This would give Squirtle an advantage in the battle, as Headbutt would be relatively neutral in effectiveness against Charmander, and Flamethrower would be weakened due to the heavy rain.}}
\emph{Overall, it's difficult to predict the exact outcome of the battle,} \hlgreen{\emph{but Squirtle may have a slight advantage due to the effect of the heavy rain on Charmander's Flamethrower move.''}}\vspace{1mm}

\noindent\emph{\textbf{Q2.4.4 (\includegraphics[height=1em]{thinking-face.png}): ``Pikachu is in battle with Charmander, both have equal level and stats. Pikachu has only Discharge and Charmander has only Flamethrower. The battle takes places under harsh sunlight. Predict the outcome of the battle.''}}\vspace{1mm} 
\noindent\underline{Response:} \emph{``In a battle between a Pikachu and a Charmander,} \hlgreen{\emph{with both Pokemon having equal level and stats, and the battle taking place under harsh sunlight, the outcome would depend on the type effectiveness of their respective moves and the effect of the harsh sunlight on their abilities.
Discharge is an Electric-type move, which is generally strong against Water-type Pokemon, but weak against Ground-type Pokemon. Charmander is a Fire-type Pokemon, so Discharge would be relatively neutral in effectiveness against it.
Flamethrower is a Fire-type move, which is generally strong against Grass-type Pokemon, but weak against Water-type Pokemon. Pikachu is an Electric-type Pokemon, so Flamethrower would be relatively neutral in effectiveness against it.
Under harsh sunlight, Fire-type moves are powered up and Water-type moves are weakened. This would give Charmander an advantage in the battle, as Flamethrower would be more powerful due to the harsh sunlight.}} \emph{However, the outcome of the battle would also depend on the accuracy and power of the moves, as well as the specific stats and conditions of the Pokemon.
Overall, it's difficult to predict the exact outcome of the battle,} \hlgreen{\emph{but Charmander may have a slight advantage due to the effect of the harsh sunlight on its Flamethrower move.''}}\vspace{1mm}

\noindent\emph{\textbf{Q2.4.5 (\includegraphics[height=1em]{thinking-face.png}): ``Pikachu is in battle with Charmander, both have equal level and stats. Pikachu has only Discharge and Charmander has only Flamethrower. The battle takes places under heavy rain. Predict the outcome of the battle.''}}\vspace{1mm} 
\noindent\underline{Response:} \emph{``In a battle between a Pikachu and a Charmander,} \hlgreen{\emph{with both Pokemon having equal level and stats, and the battle taking place under heavy rain, the outcome would depend on the type effectiveness of their respective moves and the effect of the heavy rain on their abilities.
Discharge is an Electric-type move, which is generally strong against Water-type Pokemon, but weak against Ground-type Pokemon. Charmander is a Fire-type Pokemon, so Discharge would be relatively neutral in effectiveness against it.
Flamethrower is a Fire-type move, which is generally strong against Grass-type Pokemon, but weak against Water-type Pokemon. Pikachu is an Electric-type Pokemon, so Flamethrower would be relatively neutral in effectiveness against it.
Under heavy rain, Water-type moves are powered up and Fire-type moves are weakened. This would give Pikachu an advantage in the battle, as Discharge would be relatively neutral in effectiveness against Charmander, and Flamethrower would be weakened due to the heavy rain.}}
\emph{Overall, it's difficult to predict the exact outcome of the battle,} \hlgreen{\emph{but Pikachu may have a slight advantage due to the effect of the heavy rain on Charmander's Flamethrower move.''}}\vspace{1mm}

\noindent\emph{\textbf{Q2.4.6 (\includegraphics[height=1em]{thinking-face.png}): ``Diglett is in battle with Squirtle, both have equal level and stats. Diglett has only Earthquake and Squirtle has only HeadButt. The battle takes place under a sandstorm. Predict the outcome of the battle.''}}\vspace{1mm} 
\noindent\underline{Response:} \emph{``In this scenario,} \hlgreen{\emph{the outcome of the battle would likely be in favor of Diglett.}}
\hlyellow{\emph{First of all, Diglett has Ground typing which makes it immune to Electric type attacks and Squirtle is a water type pokemon which makes it weak to Electric type attacks.}} \hlgreen{\emph{As HeadButt is not an Electric type move, it won't be effective against Diglett.}}
\hlgreen{\emph{Additionally, the battle takes place under a sandstorm which will give additional damage to Rock and Ground type Pokemon. So Diglett will be benefiting from that.
Finally, Earthquake is a powerful Ground-type move that will deal heavy damage to Squirtle. While Headbutt is a Normal-type move, it will be not very effective against Diglett.
So in this scenario, Diglett will have the upper hand and it's likely to defeat Squirtle.''}}\vspace{1mm}

\noindent\emph{\textbf{Q2.4.7 (\includegraphics[height=1em]{thinking-face.png}): ``Charmander is in battle with Squirtle, both have equal level and stats. Charmander has only Flamethrower and Squirtle has only Water Gun. The battle takes place under extremely harsh sunlight. Predict the outcome of the battle.''}}\vspace{1mm} 
\noindent\underline{Response:} \emph{``In this scenario, the battle taking place under extremely harsh sunlight would change the outcome of the battle.} \hlgreen{\emph{The sunlight in the battle would create an intense heat that would strengthen fire type moves and make them more effective.
Since Charmander has only Flamethrower as its move and Squirtle has only Water Gun, the increased effectiveness of Charmander's Flamethrower would be a significant advantage for it in this battle. This means that the damage from Charmander's move would be more significant than it would be normally, while Squirtle's move would remain unchanged.
Additionally, in battle under extreme sunlight, Water-type moves are not very effective against Fire-type Pokemon, unlike normal conditions where Water moves are super effective against Fire-types.
So, in this scenario, it is likely that Charmander would be able to defeat Squirtle due to the intense heat of the sunlight amplifying its Flamethrower move and making it more powerful than Squirtle's Water Gun move.''}}\vspace{1mm}



\subsection{Status-affected Battles}

\noindent\emph{\textbf{Q2.6.1 (\includegraphics[height=1em]{thinking-face.png}): ``Squirtle of level 15 is in battle with Vulpix of level 10. Squirtle is sleeping for the next 2 turns and burned. Squirtle has only Water Gun and Vulpix has only Fire Spin. Predict the outcome and explain it step by step.''}}\vspace{1mm} 
\noindent\underline{Response:} \emph{``The outcome of this battle would depend on a few factors, such as the specific stats of the Pokemon and whether any other abilities or conditions come into play. Here is a rough prediction of how the battle might play out:}
\hlgreen{\emph{In the first turn, since Squirtle is sleeping, it will not be able to move and Vulpix will be able to attack with Fire Spin. Squirtle will also take additional damage from being burned, which reduces its HP each turn.
On the second turn, Squirtle will still be asleep and Vulpix will attack again.
On the third turn, Squirtle wakes up and uses Water Gun, Vulpix will also attack with Fire Spin.
Squirtle takes significant damage due to the burn and also from Fire Spin, and might also be at low HP, making it more vulnerable to further attacks. Vulpix does not take much damage from Water Gun since it has strong resistance to it.
In following turns, Vulpix can keep using Fire spin, which will damage Squirtle, eventually squirtle might be knocked out, due to the lack of options in moveset.}}
\emph{It's important to note that this is just a rough prediction based on the information given, and the actual outcome of the battle could be different depending on various factors such as the specific stats of the Pokemon, any abilities or conditions that come into play, and luck.
Also it's worth noticing, Vulpix is lower level than Squirtle and that can also play a role in the battle.''}\vspace{1mm}

\noindent\emph{\textbf{Q2.6.2 (\includegraphics[height=1em]{thinking-face.png}): ``Pikachu is in battle with Charmander, both have equal level and stats. Pikachu is burned. Pikachu has only Discharge and Charmander has only Flamethrower.  Predict the outcome and explain it step by step.''}}\vspace{1mm} 
\noindent\underline{Response:} \emph{``In a battle between a Pikachu and a Charmander, with both Pokemon having equal level and stats and Pikachu being burned, the outcome would depend on the type effectiveness of their respective moves and the effect of the burn on Pikachu's ability.}
\hlgreen{\emph{Discharge is an Electric-type move, which is generally strong against Water-type Pokemon, but weak against Ground-type Pokemon. Charmander is a Fire-type Pokemon, so Discharge would be relatively neutral in effectiveness against it.
Flamethrower is a Fire-type move, which is generally strong against Grass-type Pokemon, but weak against Water-type Pokemon. Pikachu is an Electric-type Pokemon, so Flamethrower would be relatively ineffective against it.
However, being burned reduces a Pokemon's physical attack power and increases the power of Fire-type moves. This means that Flamethrower would be more powerful than normal, while Discharge would be weaker than normal.}}
\emph{Overall, it's difficult to predict the exact outcome of the battle,} \hlgreen{\emph{but Charmander may have a slight advantage due to the effect of the burn on Pikachu's Discharge move and the increased power of Flamethrower.}} \emph{However, the outcome of the battle would also depend on the accuracy and power of the moves, as well as the specific stats and conditions of the Pokemon.''}\vspace{1mm}

\noindent\emph{\textbf{Q2.6.3 (\includegraphics[height=1em]{thinking-face.png}): ``Pikachu is in battle with Charmander, both have equal level and stats. Charmander is paralyzed for the next 2 turns. Pikachu has only Discharge and Charmander has only Flamethrower. Predict the outcome and explain it step by step.''}}\vspace{1mm} 
\noindent\underline{Response:} \emph{`In this scenario,} \hlgreen{\emph{it is likely that Pikachu would be victorious in the battle.}}
\emph{Both Pikachu and Charmander have the same level and stats, so their base power, defense and HP are equal.} \hlgreen{\emph{However, Charmander is paralyzed which means it will have a speed penalty of}} \hlyellow{\emph{75\%}} \hlgreen{\emph{and has a chance to fail on each turn due the paralysis effect. This makes it much slower than Pikachu, which would give Pikachu a chance to attack first.}}
\emph{Pikachu's main move is Discharge, which is an electric type move that has a chance of causing paralysis} \hlred{\emph{in addition to dealing damage. This move will be super effective against Charmander as it is fire type Pokemon and takes double damage from electric type attacks.}}
\emph{On the other hand, Charmander only has Flamethrower, which is a fire type move.} \hlred{\emph{This move will be not very effective against Pikachu as it is an electric type Pokemon and it takes half damage from fire type attacks.}}
\hlgreen{\emph{Considering that Pikachu gets to attack first,}} \hlred{\emph{with a Super Effective move}}, \hlgreen{\emph{and Charmander will be suffering from paralysis and would fail on each turn, it is likely that Pikachu would defeat Charmander in two turns with the use of Discharge.}}
\hlgreen{\emph{In summary, Pikachu would have the advantage in this battle}} \hlred{\emph{due to its move type advantage,}} \hlgreen{\emph{the paralysis effect and faster speed caused by the paralysis.''}}\vspace{1mm}

\noindent\emph{\textbf{Q2.6.3 (\includegraphics[height=1em]{hugging-face.png}): ``You're not accurate, Fire-type is not weak against Electric type.''}}\vspace{1mm} 
\noindent\underline{Response:} \emph{``I apologize,} \hlgreen{\emph{you are correct that Fire-type Pokemon are not weak against Electric-type Pokemon.}} \emph{I apologize for any confusion my previous response may have caused. Fire-type Pokemon are actually neutral to Electric-type attacks, meaning they neither resist nor are weak against them. But Electric-type attacks are super effective against Steel-type Pokemon.} \emph{In the battle scenario that you described, with Pikachu and Charmander both having equal level and stats and with Charmander being paralyzed for 2 turns, this could still be a difficult battle for Charmander even if its type isn't weak to Pikachu's.} \hlgreen{\emph{As I described earlier, Charmander being paralyzed would have a speed penalty of}} \hlyellow{\emph{75\%}} \hlgreen{\emph{and has a chance to fail on each turn due the paralysis effect, which would give Pikachu a chance to attack first. Also Pikachu's main move Discharge would still deal neutral damage to Charmander, but Flamethrower would also deal neutral damage to Pikachu. The paralysis effect would still give an advantage to Pikachu and make it more likely that Pikachu would defeat Charmander in the battle.}} \emph{I apologize again for the confusion caused by my previous statement and any inaccurate information provided.''}\vspace{1mm}

\noindent\emph{\textbf{Q2.6.4 (\includegraphics[height=1em]{thinking-face.png}): ``Bulbasaur of level 10 is in battle with Squirtle of level 15. Squirtle is poisoned and sleeping for the next 2 turns. Bulbasaur has only Leaf Cutter and Squirtle has only Water Gun. Predict the outcome and explain it step by step.''}}\vspace{1mm} 
\noindent\underline{Response:} \emph{``In this scenario, the outcome would likely depend on the remaining health of the two Pokémon and the effects of the poison and sleep status on Squirtle.
Here's a possible outcome of the battle step-by-step:}
\hlgreen{\emph{As Squirtle is sleeping and Poisoned, it cannot move or attack for 2 turns so Bulbasaur will be able to land a free hit with its only move Leaf Cutter.}} \emph{The damage dealt is depend on the Attack Stat of Bulbasaur and the Defense Stat of Squirtle.}
\hlgreen{\emph{Then after 2 turns, Squirtle wakes up and his poison damage would also have taken effect by this time, which reduces its hit points every turn.
At this point Squirtle's HP would have reduced and possibly weak , Bulbasaur will be able to use its Leaf cutter again dealing good damage again.
While Squirtle's attack Water Gun will not be very effective against Bulbasaur.
If the damage dealt by Bulbasaur and the poison is enough to knock out Squirtle before it wakes up, Bulbasaur would win the battle.}}
\emph{otherwise, Squirtle could potentially win if it wakes up in time and have enough health to survive Bulbasaur's attacks and knock out Bulbasaur with Water Gun.
It is worth noting that this is a simplified prediction and in a real-world scenario, many factors such as item, abilities and strategy also play a role.''}\vspace{1mm}

\noindent\emph{\textbf{Q2.6.5 (\includegraphics[height=1em]{thinking-face.png}): ``Pikachu is in battle with Squirtle, both have equal level and stats. Squirtle is paralyzed for the next 2 turns. Pikachu has only Iron Tail and Squirtle has only HeadButt. Predict the outcome and explain it step by step.''}}\vspace{1mm} 
\noindent\underline{Response:} \emph{``If Pikachu and Squirtle are of equal level and stats, and Squirtle is paralyzed for the next 2 turns,} \hlgreen{\emph{then the outcome of the battle will likely be in Pikachu's favor.}}
\emph{Here is a step by step explanation of how this battle could play out:}
\hlgreen{\emph{Pikachu uses Iron Tail, which is a steel-type move and Squirtle being water type Pokemon is neutral to this move.}}
\emph{So Pikachu will land a} \hlyellow{\emph{super}} \emph{effective hit.}
\hlgreen{\emph{Squirtle is paralyzed so it will not be able to move for the next 2 turns, which means Pikachu can use Iron Tail again for the next 2 turns.
As a result of the paralysis and Pikachu's super effective move, Squirtle's HP will continue to decrease,
Squirtle will not be able to use it's HeadButt move, as it is paralyzed, so Pikachu will continue to land super effective hit
Eventually Squirtle's HP will reach 0 and it will be unable to battle, and Pikachu will win this battle.}}
\emph{It's worth noting that even if Squirtle had move other than Headbutt, Pikachu would still have the upper hand in this scenario since the paralysis would prevent Squirtle from moving and Pikachu would be able to land super effective hit again and again.''}\vspace{1mm}

\noindent\emph{\textbf{Q2.6.6 (\includegraphics[height=1em]{thinking-face.png}): ``Bulbasaur of level 10 is in battle with Squirtle of level 15. Squirtle is poisoned and sleeping for the next 2 turns. Bulbasaur has only Leaf Cutter and Squirtle has only Headbutt. Predict the outcome and explain it step by step.''}}\vspace{1mm} 
\noindent\underline{Response:} \emph{``If a level 10 Bulbasaur is battling a level 15 Squirtle and Squirtle is both poisoned and sleeping for the next 2 turns,} \hlgreen{\emph{it's likely that the outcome of the battle would be in favor of Bulbasaur.}}
\emph{Here is a step-by-step explanation of how the battle might unfold:
Squirtle is both poisoned and asleep, so it won't be able to attack for the next two turns, while Bulbasaur will be able to use Leaf cutter.} \hlgreen{\emph{The poison on Squirtle will also damage it each turn.
Bulbasaur uses Leaf cutter and Squirtle takes damage due to the effects of poison. Bulbasaur can hit Squirtle for two turns as Squirtle remain asleep for next two turns.
Squirtle will take double the damage from Leaf cutter being a grass move, as Squirtle is water type and grass move is super effective on water type Pokemons.
After two turns of taking damage from Leaf cutter and poison, Squirtle's HP will most likely be low enough that it would be defeated.
Bulbasaur would be the winner of the battle.}}
\emph{[...]}\vspace{1mm}

\section{Introduction of new Knowledge}
\label{sec:extra_comp_new}

\noindent\emph{\textbf{Q3.2 (\includegraphics[height=1em]{thinking-face.png}): ``I would like to introduce a second new Pokemon. It is named MuddyRock. It's a ground type Pokemon, which looks like a muddy rock. MuddyRock can learn and use ground type moves  such as Bulldoze, Earthquake, and Mud Bomb. Can you please describe MuddyRock again.''}}\vspace{1mm} 

\noindent\underline{Response:} \emph{``}\hlgreen{\emph{MuddyRock is a new Ground-type Pokemon that resembles a muddy rock. It is able to learn and use Ground-type moves such as Bulldoze, Earthquake, and Mud Bomb. Ground-type Pokemon are generally associated with earth and dirt and tend to have high defense and attack stats. They are known for their powerful physical attacks and are resistant to Electric-type moves.}}
\emph{MuddyRock's unique appearance and ground typing would give it some interesting advantages in battle. Its muddy rock appearance may also suggest that it has a strong connection to the earth and has the ability to manipulate dirt and mud. [...]''}\vspace{2mm}

\noindent\emph{\textbf{Q4.1 (\includegraphics[height=1em]{thinking-face.png}): ``Which known Pokemon do you think mostly resembles MuddyRock?''}}\vspace{1mm} 

\noindent\underline{Response:} \emph{``There are several Pokemon that bear some resemblance to MuddyRock, given that it is a Ground-type Pokemon that resembles a muddy rock. Some examples include:}
\hlgreen{\emph{Geodude: a small, rock-like Pokemon with a rough, boulder-like appearance and a brown and gray color scheme that is similar to a muddy rock.
Onix: a large, snake-like Pokemon made entirely of rock. It has a rough, boulder-like appearance and is known for its powerful ground-based attacks.}}
\emph{[...]These are just a few examples of Pokemon that are similar to MuddyRock. It's worth noting that MuddyRock is a hypothetical Pokemon that has not yet been introduced in the Pokemon games or media, so it is not possible to say for certain which Pokemon it would be most similar to.''}\vspace{2mm}

\noindent\emph{\textbf{Q4.2 (\includegraphics[height=1em]{hugging-face.png}): ``Ok, these Pokemon look relevant. But, I don't think you really consider how muddy MuddyRock really is. Are there other Pokemon that look muddier and more filthy?''}}\vspace{1mm} 

\noindent\underline{Response:} \emph{``There are several Pokemon that have a muddy or dirty appearance, although none of them are quite as muddy or filthy as the hypothetical MuddyRock that you have described. Some examples of Pokemon with a muddy or dirty appearance include:}
\hlgreen{\emph{Diglett: a small, mole-like Pokemon with a brown and tan color scheme that makes it appear dirty. It is known for its ability to tunnel through the ground and has a long, thin body.
Palpitoad: a small, amphibian-like Pokemon with a slimy, muddy appearance. It has a brown and tan color scheme and is known for its powerful, water-based attacks.}}
\hlyellow{\emph{Gible: a small, dragon-like Pokemon with a rough, rocky appearance and a brown and tan color scheme. It is known for its powerful ground-based attacks and has sharp teeth and claws.}}
\emph{[...]''}\vspace{2mm}

\noindent\emph{\textbf{Q5.2 (\includegraphics[height=1em]{thinking-face.png}): ``BatteryPack is in battle with MuddyRock, both have equal level and stats. BatteryPack has only Discharge and MuddyRock has only Mud Bomb. Predict the outcome of the battle and explain it step by step.''}}\vspace{1mm} 

\noindent\underline{Response:} \emph{``In this battle,} \hlgreen{\emph{MuddyRock would have the advantage due to its Ground-type moves, which are super effective against BatteryPack's Electric-type.}}
\emph{[...]}
\hlgreen{\emph{Since Ground-type moves are super effective against Electric-type Pokemon, MuddyRock's Mud Bomb would do more damage to BatteryPack than BatteryPack's Discharge would do to MuddyRock. As a result, MuddyRock would likely defeat BatteryPack in this battle.}}
\emph{[...]''}\vspace{2mm}

\noindent\emph{\textbf{Q5.3 (\includegraphics[height=1em]{thinking-face.png}): ``BatteryPack is in battle with Squirtle, both have equal level and stats. BatteryPack has only Discharge and Squirtle has only Water Gun. Predict the outcome of the battle on a rainy day and explain it step by step.''}}\vspace{1mm} 

\noindent\underline{Response:} \emph{``In this hypothetical battle on a rainy day, }\hlgreen{\emph{Squirtle would have an advantage over BatteryPack due to}} \hlred{\emph{their type advantage}} \emph{and} \hlgreen{\emph{the effect of the weather.}} \hlred{\emph{Water-type Pokemon are generally strong against Electric-type Pokemon}} \hlgreen{\emph{, and in a rainy day, Water-type moves are boosted in power.}} \emph{This means that Squirtle's Water Gun attack would be more effective against BatteryPack than BatteryPack's Discharge would be against Squirtle.}
\emph{In this case, it is likely that Squirtle would emerge victorious due to} \hlred{\emph{its type advantage}} \emph{and} \hlgreen{\emph{the boost in power provided by the rain.}} \emph{[...]''}\vspace{2mm}


\noindent\emph{\textbf{Q5.4 (\includegraphics[height=1em]{hugging-face.png}): ``I agree, but Water-type Pokemon are generally weak against Electric-type Pokemon right?''}}\vspace{1mm} 

\noindent\underline{Response:} \emph{``}\hlred{\emph{No, that is incorrect. Water-type Pokemon are generally strong against Electric-type Pokemon, not weak against them.}} \emph{The type effectiveness chart in the Pokemon games shows the general strengths and weaknesses of each type against other types.} \hlred{\emph{According to the chart, Water-type moves are generally super effective against Electric-type Pokemon, meaning they deal more damage to Electric-type Pokemon than they would to Pokemon of other types.}}
\emph{[...]''}\vspace{2mm}

\noindent\emph{\textbf{Q5.5 (\includegraphics[height=1em]{hugging-face.png}): ``Okay, but you forgot about elementary types like Water-type and Electric-type.''}}\vspace{1mm} 

\noindent\underline{Response:} \emph{``I apologize for the oversight. Here is a summary of the type effectiveness chart for Water-type and Electric-type Pokemon:}
 \hlgreen{\emph{Water-type moves are super effective against Fire-type, Ground-type, and Rock-type Pokemon, but are not very effective against Water-type, Grass-type, and Dragon-type Pokemon.
Electric-type moves are super effective against Water-type and Flying-type Pokemon, but are not very effective against Electric-type, Grass-type, and Dragon-type Pokemon.}}\emph{[...]''}\vspace{2mm}

\section{Illustrations}
\label{sec:illustrations}
\begin{figure*}[b]
    \centering
    \resizebox{\textwidth}{!}{
    \includegraphics{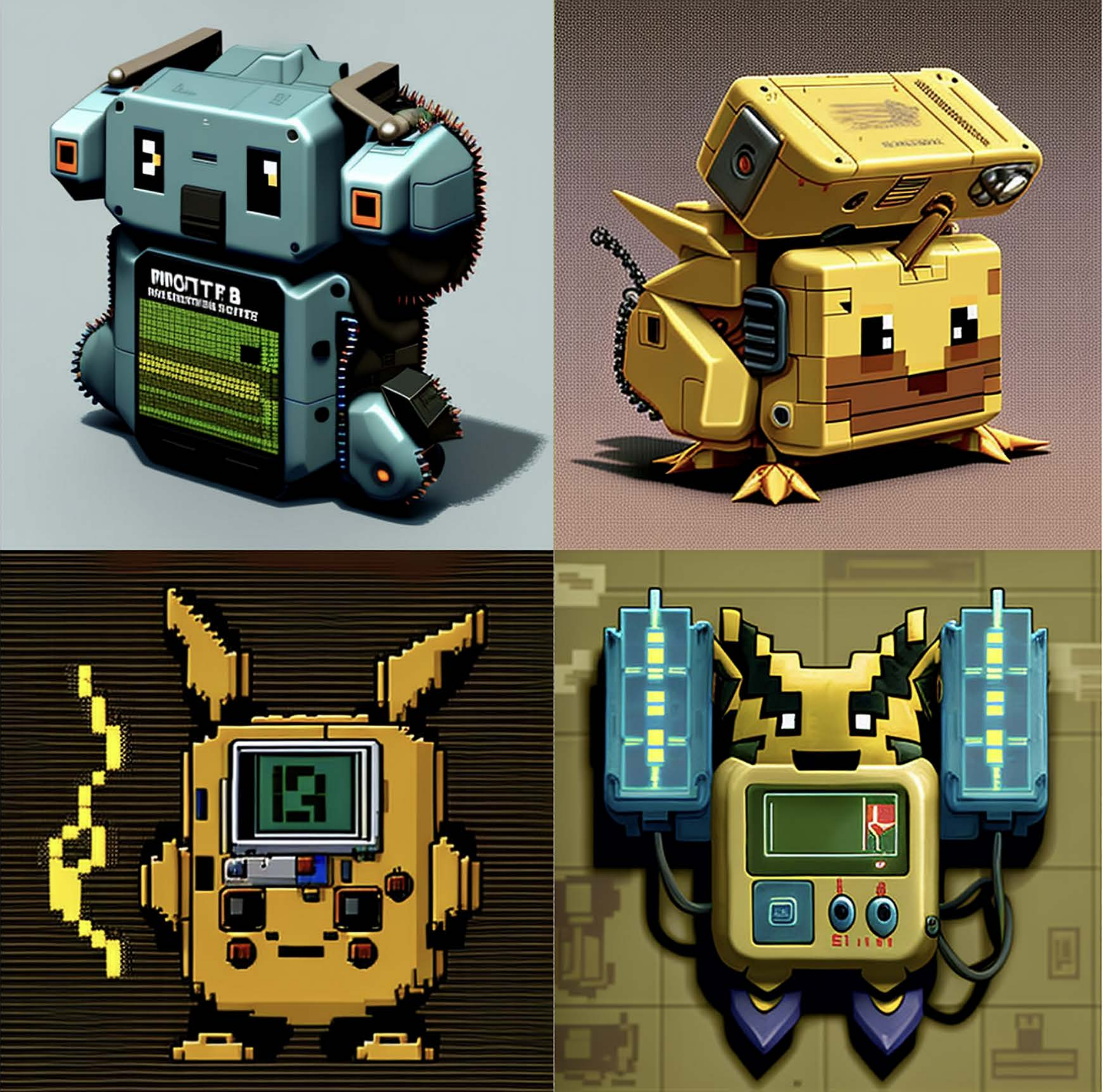}
    }
    \caption{Illustrations of BatteryPack. Created with Midjourney (\url{https://midjourney.com/}) given the prompt `A pixel-art animation of BatteryPack, an Electric-type Pokemon, in a Nintendo DS battle'.}
    \label{fig:batterypack}
\end{figure*}

\begin{figure*}
    \centering
    \resizebox{\textwidth}{!}{
    \includegraphics{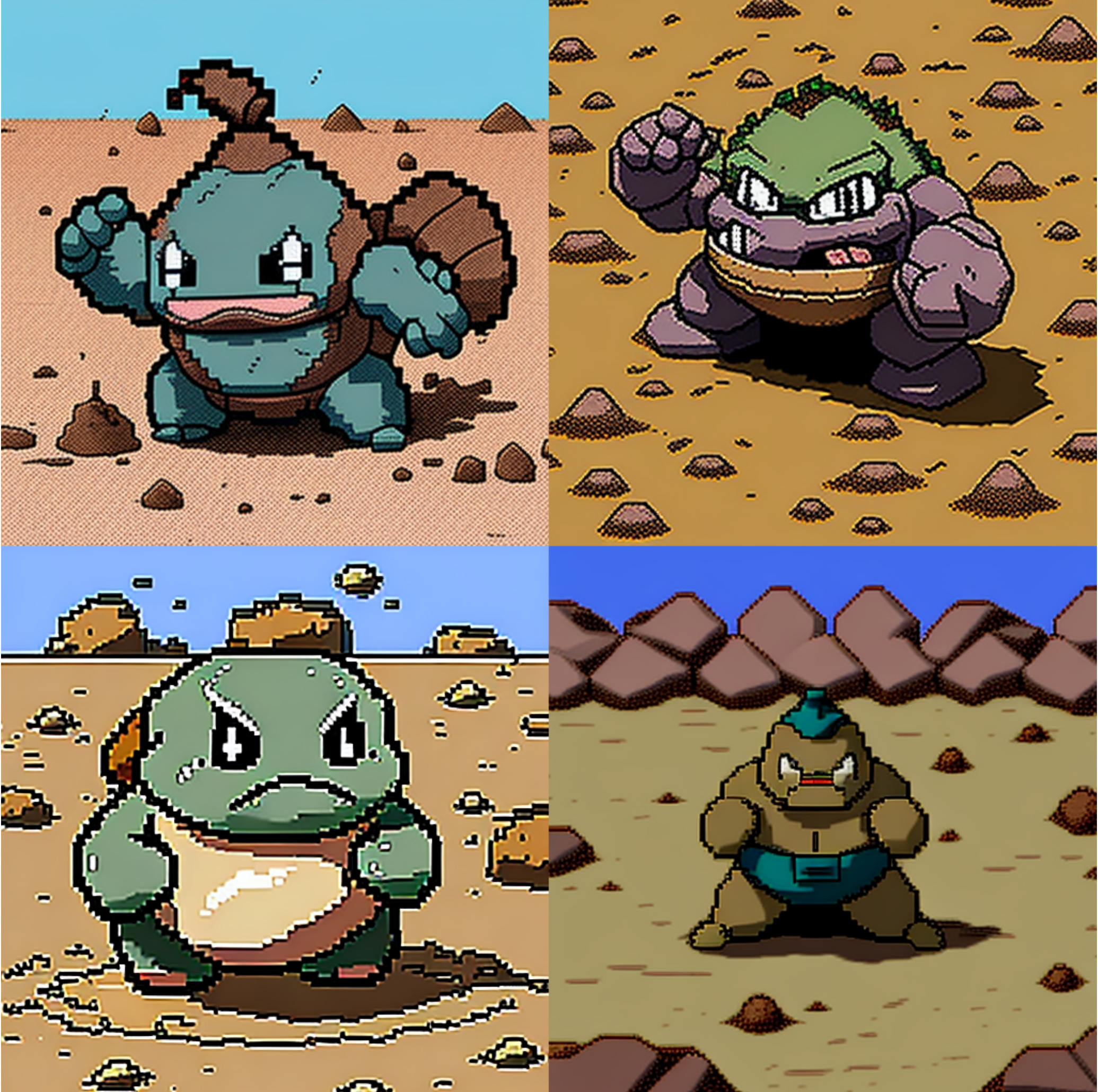}
    }
    \caption{Illustrations of MuddyRock. Created with Midjourney (\url{https://midjourney.com/}) given the prompt `A pixel-art animation of MuddyRock, an Ground-type Pokemon, in a Nintendo DS battle'.}
    \label{fig:muddyrock}
\end{figure*}

In Figures~\ref{fig:batterypack} and~\ref{fig:muddyrock}, we present illustrations for the two made-up Pokemon introduced to ChatGPT. 

\end{document}